\documentclass[10pt,twocolumn,letterpaper]{article}
\PassOptionsToPackage{table,dvipsnames}{xcolor}
\usepackage{iccv}              

\usepackage{multirow} 
\usepackage{pifont}
\usepackage{amsmath}
\usepackage{booktabs}
\usepackage{array}
\newcommand{\corresponding}{\footnotemark[1]}

\definecolor{iccvblue}{rgb}{0.21,0.49,0.74}
\usepackage[pagebackref,breaklinks,colorlinks,allcolors=iccvblue]{hyperref}

\def\paperID{7555} 
\def\confName{ICCV}
\def\confYear{2025}

\title{MMCR: Advancing Visual Language Model in Multimodal Multi-Turn Contextual Reasoning}

\author{
Dawei Yan\textsuperscript{\rm 1,2},
Yang Li\textsuperscript{\rm 2},
Qingguo Chen\textsuperscript{\rm 2},
Weihua Luo\textsuperscript{\rm 2},
Peng Wang\textsuperscript{\rm 1},\\
Haokui Zhang\textsuperscript{\rm 1 \corresponding},
Chunhua Shen\textsuperscript{\rm 3}\\
\textsuperscript{\rm 1}School of Cybersecurity, Northwestern Polytechnical University \quad
\textsuperscript{\rm 2}AI Business, Alibaba Group \quad \\
\textsuperscript{\rm 3} College of Computer Science and Technology, Zhejiang University\\
\and
}

\begin{document}
\maketitle
\renewcommand{\thefootnote}{\fnsymbol{footnote}}
\begin{NoHyper}
\footnotetext[1]{Corresponding author.}
\end{NoHyper}
\renewcommand{\thefootnote}{\arabic{footnote}}
\begin{abstract}



Compared to single-turn dialogue, multi-turn dialogue involving multiple images better aligns with the needs of real-world human-AI interactions. Additionally, as training data, it provides richer contextual reasoning information, thereby guiding the model to achieve better performance. However, existing vision-language models (VLMs) primarily rely on single-turn dialogue training and evaluation benchmarks. In this paper, following the characteristics of human dialogue, such as focused topics and concise, clear content, we present MMCR (Multimodal Multi-turn Contextual Reasoning), a novel dataset comprising: (1) MMCR-310k- the largest multi-image multi-turn instruction tuning dataset with 310K contextual dialogues, each covering 1-4 images and 4 or 8 dialogue turns; and (2) MMCR-Bench - a diagnostic benchmark featuring dialogues, spanning 8 domains (Humanities, Natural, Science, Education, etc.) and 40 sub-topics. Extensive evaluations demonstrate that models fine-tuned with MMCR-310k achieve 5.2\% higher contextual accuracy on MMCR-Bench, while showing consistent improvements on existing benchmarks (+1.1\% on AI2D, +1.2\% on MMMU and MMVet). MMCR and prompt engineering will be released publicly.


\end{abstract}

\section{Introduction}
\label{sec:intro}

\begin{figure*}[t]
    \centering
    \includegraphics[scale=0.175]{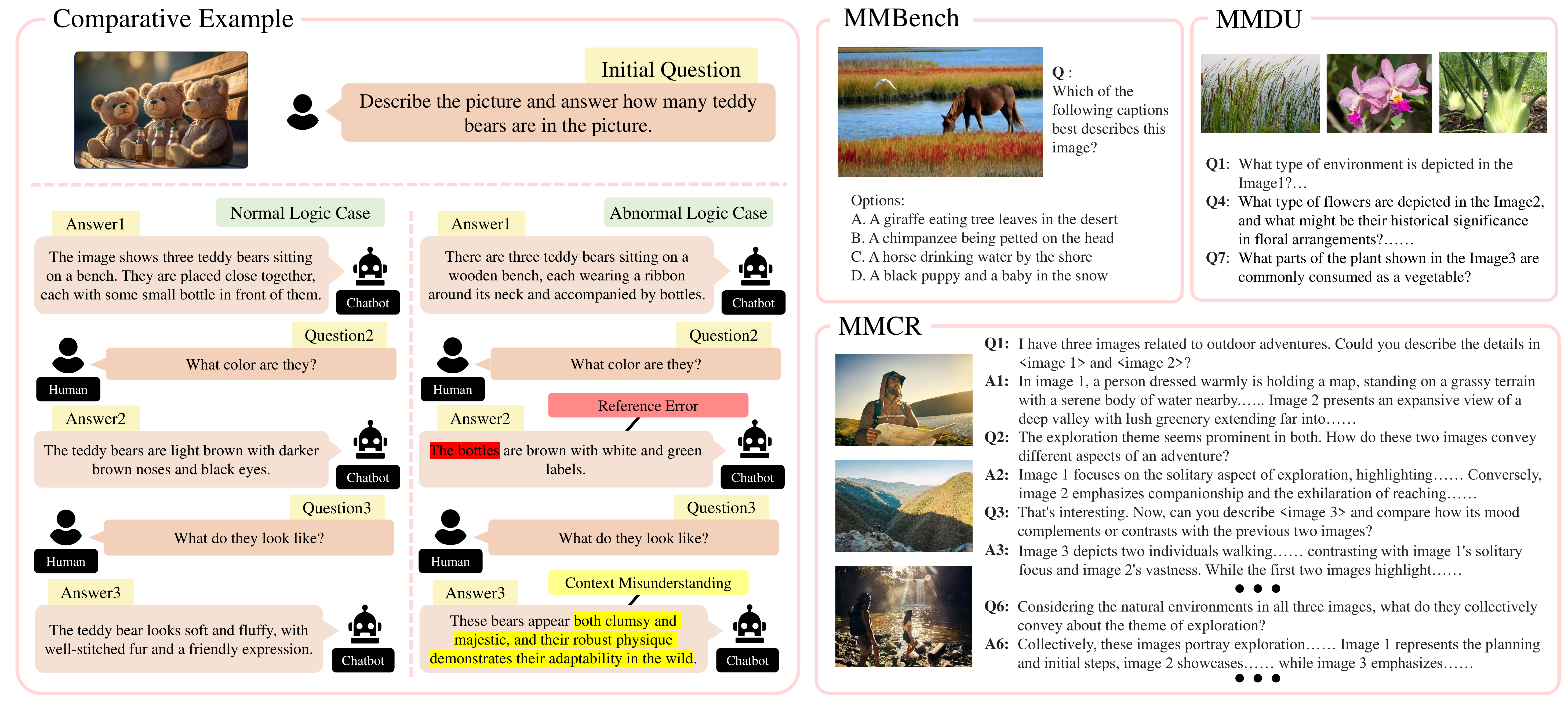}
    \caption{Samples comparison. On the left side: two examples corresponding to a good contextual reasoning sample and a bad contextual reasoning sample with errors during multi-turn dialogue. On the right side: a single-image, single-turn dialogue sample from MMBench \cite{liu2024mmbench}, a multi-turn, multi-image dialogue sample from MMDU \cite{liu2025mmdu}, and a sample from our proposed MMCR.}
    \label{fig:dataset_comp}
\end{figure*}

The pursuit of establishing Artificial General Intelligence (AGI) capable of providing expert-level responses has long been a pressing goal within the academic community. Such mature agents are expected to engage in multi-turn interactions with human users, process long-context dialogue information, and demonstrate multi-turn instruction following ability, which are also increasingly common requirements in real-world scenarios.

In response to the current state, numerous outstanding studies and benchmarks in the domain of Large Language Models (LLMs) have been developed to enhance their ability to handle long-context multi-turn dialogue scenarios, including but not limited to sparse and efficient attention mechanisms \cite{dao2023flashattention,liu2022dynamic}, position encoding extrapolation and dynamic adjustment \cite{ma2025mesa}, context summarization and memory bank construction, as well as various multi-turn dialogue fine-tuning datasets and evaluation benchmarks \cite{bai2024mt}. However, for open-source VLMs, the mainstream training data and evaluation benchmarks \cite{liu2024mmbench} are largely confined to handling single-image, single-interaction modes. However, in real-world human-AI interactions, multi-turn dialogues involving multiple images better align with human conversational habits and practical application needs. Furthermore, multi-turn and multi-image dialogue datasets contain richer contextual reasoning information, which significantly aids in enhancing the performance of data-driven VLMs.

To overcome this problem, some researchers have attempted to address this issue from the perspective of building a data foundation \cite{cui2020mutual, bai2024mt}. However, most of these efforts focus solely on dialogue content. Very recently, MMDU \cite{liu2025mmdu} has constructed high-quality training data and benchmarks using GPT-4o based on Wikipedia to enhance and evaluate VLMs in such scenarios, but it does not emphasize strengthening contextual logic, maintaining cross-turn dialogue consistency, or accurately parsing long-range contextual dependencies. These capabilities are essential for deploying AGI in real-world applications involving continuous multimodal inputs. Moreover, logical reasoning with long-contexts also contributes to enhancing the model's performance on public benchmarks. Therefore, there is an urgent need in academia to develop a multi-dimensional training data and evaluation framework that includes complex multi-turn dialogues, strong contextual associations, and dynamic scene adaptation. 

Building on this, we introduce MMCR (Multimodal Multi-turn Contextual Reasoning), which includes MMCR-310k and MMCR-Bench, a multi-turn instruction fine-tuning dataset with a mix of single-image and multi-image interactions, and an evaluation benchmark. This instruction fine-tuning dataset and evaluation benchmark are constructed based on the multimodal text-image interleaved dataset OmniCorpus \cite{li2024omnicorpus}, though GPT-4o \cite{openai2024gpt4o} with the guidance of carefully designed prompt engineering. The generated multi-turn dialogue datasets, which feature strong contextual logic and simulate real-world user-agent interactions, also conduct evaluations on specific dimensions of multi-turn dialogue contextual reasoning.

Specifically, MMCR features the following characteristics: (1) Multi-turn Interaction and Realism: MMCR-310k contains 210k single-image and 100k multi-image multi-turn dialogue data, with dialogues focused on images over four and eight rounds, respectively, aiming to simulate real-world user-chatbot interactions. (2) Strong Contextual Relevance and Logical Progression: We emphasize that each round of dialogue should progressively delve into image details, inter-image relationships, and related themes, while maintaining a clear contextual association. Subsequent questions must build upon prior requests or responses. (3) Evaluation Openness and Extensiveness: We utilize multi-level annotators provided by \cite{ridnik2021imagenet} to label images from OmniCorpus \cite{li2024omnicorpus}, selecting labeled samples from eight major domains—Humanities, Nature, Education, Science, Mechanics, Animals, Plants, and Architecture—across 40 sub-topics for the evaluation set. Using GPT-4o as the evaluator, we score the model's responses across five dimensions: Precision and Conciseness, Consistency of Contextual References, Logical Contextual Relationship, Clarity of Dialogue Theme, and Absence of Redundancy. This comprehensive evaluation assesses the model's performance in long-context multi-turn dialogues. Several samples from difference datasets are compared in Fig. \ref{fig:dataset_comp}


We validate MMCR using Ovis \cite{lu2024ovis}, a state-of-the-art (SOTA) VLM architecture that innovatively employs a learnable visual embedding table for visual encoding. After instruction fine-tuning with MMCR, Ovis achieved obviously improvement on MMCR-Bench, which revealed the limitations of existing models in logical contextual reasoning in multi-turn dialogues, and validated the necessity of constructing such a dataset. Besides, incorporating our proposed MMCR into the supervised fine-tuning stage further enhances performance on existing benchmarks. For instance, building on Ovis's significant outperformance over other open-source models—even surpassing GPT-4o on some metrics—it achieves additional improvements on multiple benchmarks such as AI2D \cite{kembhavi2016diagram}, MMMU \cite{yue2024mmmu}, and MMV \cite{yu2023mm}.


Furthermore, based on our detailed analysis of the experimental results, we identify a subtle yet critical point to consider when training large models, which we refer to as the ``less is more" phenomenon. During the high-quality supervised fine-tuning phase, simply adding more data does not always lead to better performance. Maintaining a balanced proportion of data across different task types is equally important. In fact, to preserve this balance, reducing the amount of data for certain task types can sometimes yield better results. This phenomenon is particularly pronounced for models with fewer parameters. This finding provides valuable insights for subsequent large model training, particularly in supervised fine-tuning.

In summary, the major contributions are: 

\begin{itemize}
\item We propose MMCR, a multimodal collection featuring a hybrid instruction-tuning dataset (310K dialogues with 4-8 turns) and a GPT-4o-grounded benchmark, specifically designed to enhance and assess the multi-turn dialogue contextual reasoning ability of VLMs in real-world human-AI dialogues.


\item Systematic evaluation shows that MMCR enables precise and effective improvements and assessments in the context of multimodal multi-turn dialogues. It also shows enhanced performance on existing benchmarks.

\item We discover the ``Less is More" phenomenon in supervised fine-tuning of large models and demonstrated it through experiments. This phenomenon challenges the conventional belief that ``more data is always better" and emphasizes the importance of data distribution in addition to data volume. It provides valuable insights for future large model training efforts.

\end{itemize}
\section{Related Work}
\subsection{Multi-turn Instruction Tuning Datasets}
Recent advancements in instruction tuning \cite{liu2023visual} have significantly enhanced model capabilities across various downstream tasks. \cite{li2023textbind} introduced a framework that requires minimal annotation to enable LLMs to follow multi-turn multimodal instructions, requiring only image-description pairs to generate multi-turn multimodal instruction-response dialogues from a language model. Niu et al.\cite{niu2024multimodal} introduced a new multimodal, multi-turn posture detection dataset, MmMtCSD, and proposed a novel multimodal large language model framework, MLLM-SD, aimed at advancing the practical application of posture detection research. Maheshwary \cite{maheshwary2024m2lingual} proposed a taxonomy-guided, case-based generation of a multilingual, multi-turn instruction fine-tuning dataset, M2Lingual, containing 182K total instruction fine-tuning data, covering 70 languages and 17+ NLP tasks. To further align LLM outputs with human preferences, R2S \cite{hou2024raw} used dialogue chain logic to guide LLMs in generating knowledge-intensive multi-turn dialogues for instruction fine-tuning, and constructs the K-BENCH and GINSTRUCT instruction datasets using open-source datasets and domain-specific knowledge documents. MMDU \cite{liu2025mmdu}, starting from the current status of existing VLM benchmarks that primarily focus on multiple-choice questions or short-format responses, clustered related images and textual descriptions from the open-source Wikipedia, and constructed question-answer pairs with the help of human annotators using GPT-4o. They proposed the MMDU-45k instruction fine-tuning set and the MMDU evaluation benchmark, paving the way to bridge the gap between current VLM models and practical application needs.

Multi-turn dialogue has garnered significant attention from scholars in the field, leading to the creation of numerous multi-turn dialogue instruction-following datasets. However, existing efforts primarily focus on the linguistic level and are confined to specific knowledge domains, with relatively few datasets addressing multi-turn dialogues involving multiple images. Additionally, current datasets often overlook issues such as logical coherence and clear referencing within the context. The constructed datasets resemble a patchwork of single-turn dialogues rather than a coherent, topic-centered multi-turn conversation.



\subsection{VLM Benchmarks}
The rapid development of the multimodal field has driven the industry to continuously propose evaluations of the capabilities of the models in various tasks and domains. Nowadays, the academic community has developed numerous high-quality evaluation benchmarks~\cite{lu2022learnexplainmultimodalreasoning,yue2024mmmu,liu2023mmbench,liu2023visualinstructiontuning}, establishing a standardized and objective evaluation system. However, the evaluation of visual language models has traditionally focused on atomic tasks such as visual question answering~\cite{fu2024mmecomprehensiveevaluationbenchmark}, image understanding and captions~\cite{hiippala2021ai2d}, OCR recognition~\cite{Liu_2024}, and diagram analysis~\cite{masry2022chartqabenchmarkquestionanswering}, where single-instance understanding is prioritized over contextual coherence. While current benchmarks are effective in measuring basic perceptual capabilities, they fall short in capturing the contextual logical coherence and dynamic nature of real-world dialogues. In such realistic, context-driven dialogues, conversations between humans and chatbots should exhibit strong contextual logic, concise and clear content, and a clear storyline and theme. This necessitates that the model possess the ability to reference multiple images and integrate iterative knowledge. To address this, we design and develop MMCR, aiming at enhancing the aforementioned capabilities of multimodal models through carefully crafted multi-turn dialogue context reasoning instruction fine-tuning data. Additionally, MMCR evaluates whether the model can handle strong contextual logical relationships to further meet the need for expert-level AGI interaction in real-world scenarios.

\begin{figure*}[t]
    \centering
    \includegraphics[scale=0.19]{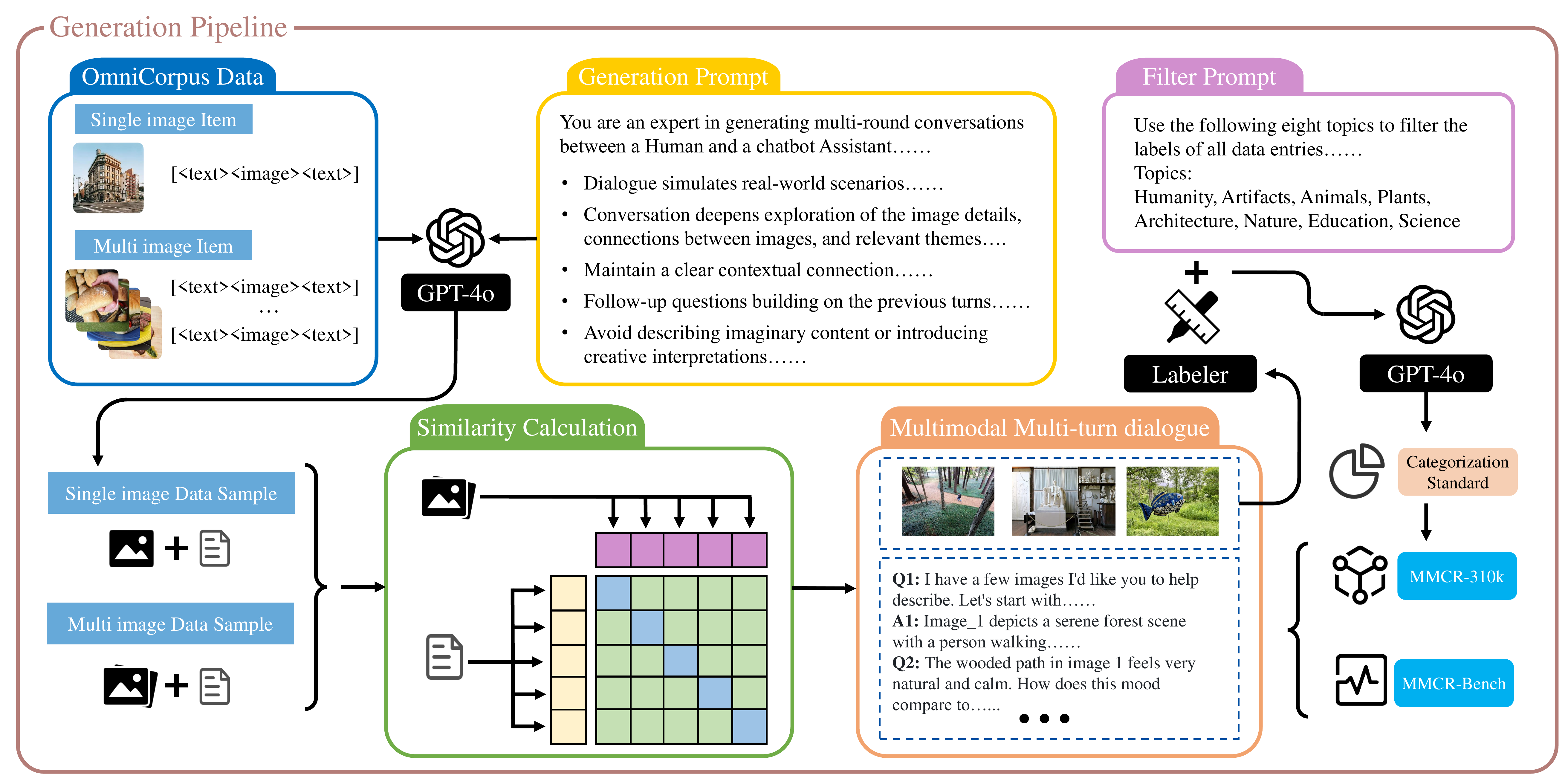}
    \caption{The pipeline of constructing MMCR. First, we randomly select 1.6 million samples from OmniCorpus as base samples. Then, we generate dialogue content using GPT-4o guided by our carefully designed prompt engineering. Next, we filter the constructed dataset using CLIP based on the semantic similarity between images and their corresponding dialogues, resulting in 310k high-quality samples. Finally, we analyze the topics of the obtained samples using an automatic labeler and randomly select samples from 40 topics based on the statistical information to build MMCR-Bench.}
    \label{fig:pipeline}
\end{figure*}

\section{MMCR}

\subsection{Overview}
In both the NLP and multimodal fields, researchers have long been eager to expand the context window length of models. A widely accepted notion is that a longer context window means the ability to receive more extended context, allowing for better information integration and providing more reasonable responses. However, despite significant progress in extending the context window length of VLMs, their practical performance in real-world multi-turn dialogues remains limited, particularly when handling interleaved multi-image inputs in long conversation contexts \cite{liu2025mmdu}. More importantly, evaluating a model's interactive responses in multi-turn dialogue settings should not be confined to the number of dialogue turns, the length of dialogue, or the number of input images. Instead, it depends on whether the model can demonstrate excellent contextual logic reasoning capabilities, effectively distilling key information from multi-turn dialogues and making reasonable conclusions.

To further enhance and evaluate the long-context reasoning abilities of VLMs in multimodal, multi-turn dialogue scenarios, we propose MMCR, a dataset that includes a multimodal multi-turn dialogue instruction fine-tuning dataset and corresponding evaluation benchmarks. It consists of MMCR-310k, which includes 210k single-image and 100k multi-image high-quality multi-turn dialogue data with strong contextual logic relationships, and MMCR-Bench, an evaluation benchmark containing 600 carefully selected single/multi-image mixed evaluation data with multi-level annotations. We not only break through the limitations of previous evaluation benchmarks, which typically involve only single images, brief question-answer pairs, and fewer dialogue turns, but also innovatively introduce the concept of strong contextual logic through prompt engineering in the dataset construction, ensuring that the constructed multi-turn dialogue data has clear contextual logic relationships, concise responses, and a well-defined theme. In Sections 3.2 and 3.3, we will provide detailed descriptions of MMCR-310k and MMCR-Bench, including the construction pipeline for the instruction fine-tuning data, the specific distribution, as well as the annotation and construction of the evaluation set and the corresponding objective evaluation process.

\subsection{MMCR-310K}
\textbf{Data Collection.} MMCR aims to provide VLMs with multi-turn multimodal dialogue data that has strong contextual relevance and logical reasoning, enhancing and evaluating the performance of existing models in real-world human-AGI interaction scenarios. To construct multi-turn multimodal dialogues with coherent contextual logic, ensuring that the generated images and text are highly consistent is essential. If there is no good correspondence between the reference images or between the images and text, the generated data will become too random, causing jumps between dialogue turns, disrupting contextual logic, and ultimately leading to responses that do not meet expectations.
OmniCorpus~\cite{li2024omnicorpus} is a massive multimodal dataset containing billions of image-text pairs, with its open-source version, OmniCorpus-CC-210M, covering a wide range of languages and scenarios from simple to complex. The dataset collects and extracts large amounts of high-quality documents from the internet using an efficient data engine and goes through multi-stage filtering, ultimately constructing multimodal data in formats like image-text interleaving.
The image-text interleaved data it generates perfectly aligns with the homogeneity we pursue, with high relevance between images and between images and text. Therefore, when generating data, we do not need secondary clustering or other operations to maintain data relevance. Based on the above analysis, we sampled some data from OmniCorpus as the foundation for constructing our dataset.

\begin{table}[t]
\centering
\resizebox{0.47\textwidth}{!}{
\begin{tabular}{lcc}
\hline
\multicolumn{1}{c}{Statistic}                      & \multicolumn{2}{c}{Number}         \\ \hline
\textbf{MMCR-310k}  & 210k single & 100k multi           \\
- Avg./Max. Image\&Text Tokens & 1.4k/1.8k   & 3.5k/6k              \\
-Images number                 & 1           & 2 / 3 / 4                \\
-Proportion                    & -           & 61.5\%/27.6\%/10.9\% \\
-Turns                         & 4           & 8                    \\
-Total image                   & 210k        & 270k                 \\
-Number of QA Pairs            & 840k &   870k        \\ \hline
\textbf{MMCR-Bench}                    & 400         & 200                  \\
-Proportion                    & -      & 50.5\%/30.5\%/19\%              \\
-Total image                   & 400         & 537                  \\
-Number of QA Pairs            &  1.6k & 1.6k        \\ \hline
\end{tabular}
}
\caption{Statistics of MMCR-310k and MMCR-Bench.}
\label{tab:data_statistic}
\end{table}

\noindent \textbf{Generation Pipeline.} With a high-quality source of image-text data, we carefully craft prompts to guide GPT-4o in generating the multimodal multi-turn dialogue data for MMCR based on the provided data. We conduct complex prompt engineering to ensure the generation of multi-turn dialogue data with strong contextual logic. The key focus is: ``Ensure that each turn of the dialogue progressively deepens the exploration of image details, connections between images, and related topics. The dialogue must maintain clear contextual continuity, with subsequent questions based on previous requests or questions and their corresponding answers. At the same time, the AI assistant's responses must be detailed and contextually relevant, focusing only on observable details directly inferred from the images, avoiding descriptions of fictional content or introducing creative interpretations." This is the key to constructing multi-turn dialogues with contextual logic. At the same time, we strictly limit the dialogue style in the prompts. We stipulate that the special token marking for each image only appears once at the beginning of the first dialogue where the image is mentioned, such as “Image\_n: $\textlangle$image$\textrangle$,” where n represents the sequence number of the image in the current case. We extract 1.2M single-image samples and 400k multi-image samples from OmniCorpus-CC-210M and combined them with fine-grained prompts to deliver to GPT-4o for multi-turn dialogue data generation. 

The resulting samples are then strictly filtered. We use the classic image-text encoder, CLIP~\cite{radford2021learningtransferablevisualmodels}, as a referee, comparing the semantic similarity between images and their corresponding dialogues in the generated results. Using strict regular functions, we filter out entries with missing images, formatting errors, unclear text, and mismatches between images and text, ultimately obtaining 210K single-image samples and 100K multi-image samples. The complete construction pipeline is shown in Figure \ref{fig:pipeline}.

\begin{figure}[t]
    \centering
    \includegraphics[scale=0.35]{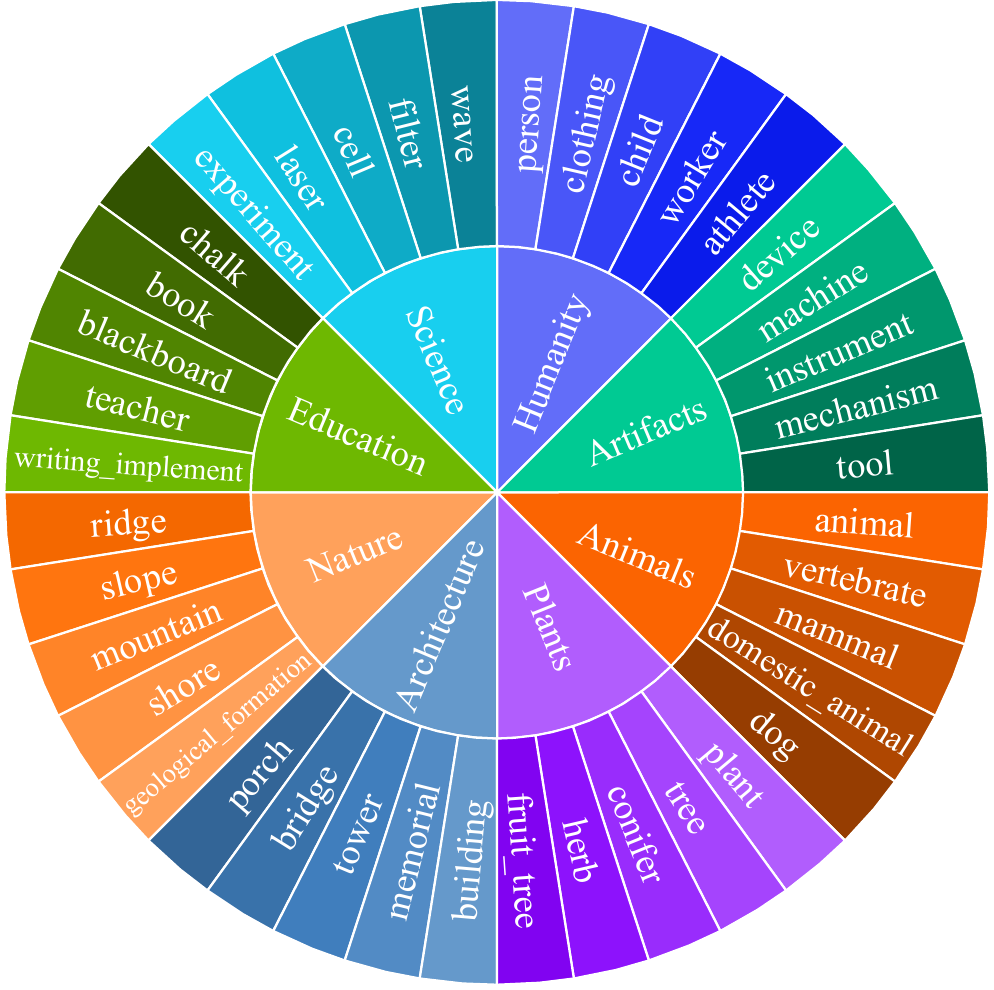}
    \caption{Topic distribution of MMDU-Bench.}
    \label{fig:topic_distribution}
\end{figure}

\noindent \textbf{Note}: During the data construction process, we find that when generating data based on a single image and the corresponding image-text interleaved text, the resulting data often exhibited hallucination issues. We believe this is due to the limited amount of semantic information contained in a single image. Relying solely on a single image makes it difficult to generate multi-turn dialogues with coherent contextual logic that closely align with reality. As a result, GPT-4o tends to engage in artistic creation during dialogue generation, introducing content that does not belong to the image-text interleaved data. Therefore, we restrict the number of dialogue turns generated for single-image data to 4 turns and allow up to 8 turns for multi-image data.

\noindent \textbf{Data Statistics.} The proposed MMCR dataset contains a total of 310k single/multi-image mixed multi-turn instruction fine-tuning dialogue data. Tab.~\ref{tab:data_statistic} shows the specific distribution of our data. Among them, the number of dialogue turns for single-image data is 4, with an average image-text mixed token length of 1.4k and a maximum of 1.8k. For multi-image data, the number of dialogue turns is 8, with an average image-text mixed token length of 3.5k and a maximum of 6k. Each entry in the multi-image data contains 2 to 4 images, and the distribution of different image quantities is 61.5\%/27.6\%/10.9\%. All data follows a strict alternating human-AI assistant dialogue mode, with each dialogue turn closely related to the image. It is important to emphasize that MMCR focuses on presenting multi-turn multimodal dialogues with strong contextual relevance, clear logic, and themes. Therefore, designing data with a large number of images or extending the length of responses in dialogues is not our main goal. We believe that a moderate length and number of images provide the best setting to enhance and evaluate the contextual reasoning capabilities of VLMs.

\subsection{MMCR-Benchmark}

\textbf{Data Annotation and Construction.} To conduct a comprehensive evaluation, we aim for the constructed evaluation benchmark to cover as many fields and topics as possible, in order to assess the VLMs’ overall knowledge base and comprehension abilities. Therefore, we annotate the data based on the images and set up multiple topics for screening to build an evaluation benchmark with evenly distributed categories and broad coverage, namely MMCR-Bench.

We first use the labeler provided by \cite{ridnik2021imagenet} to annotate the images in the data samples. The pre-trained model of this labeler has been efficiently trained on ImageNet-21k and possesses robust labeling capabilities. The labeler can provide multi-level tags, so we aggregate all the labels from the entire dataset and count the frequency of each tag across all images, before passing the results to GPT-4o for further screening. We have defined a total of eight major categories: humanity, artifacts, animals, plants, architecture, nature, education, and science, and we use GPT-4o to partition the aggregated tag statistics according to these categories. For each major category, the top five most frequent tags are selected as subtopics, resulting in a final set of 40 tags with the highest occurrence across these categories that serve as the coverage categories for MMCR-Bench. Subsequently, data entries that include images with the corresponding tags are considered to belong to the respective topics. Within each major category, we randomly select 10 samples from the single-image data and 5 from the multi-image data, ultimately obtaining a total of 600 multimodal multi-turn dialogue entries to serve as the final MMCR-Bench evaluation benchmark. We show the distribution of these topics in Fig.~\ref{fig:topic_distribution}. Thanks to the high-precision labeler and GPT-4o’s screening, our constructed MMCR-Bench evaluation benchmark offers very wide coverage, encompassing various aspects of daily life. Not only does it test the contextual reasoning ability of the model, but it also extensively examines whether the model possesses sufficient general knowledge.

\begin{figure}[t]
    \centering
    \includegraphics[scale=0.17]{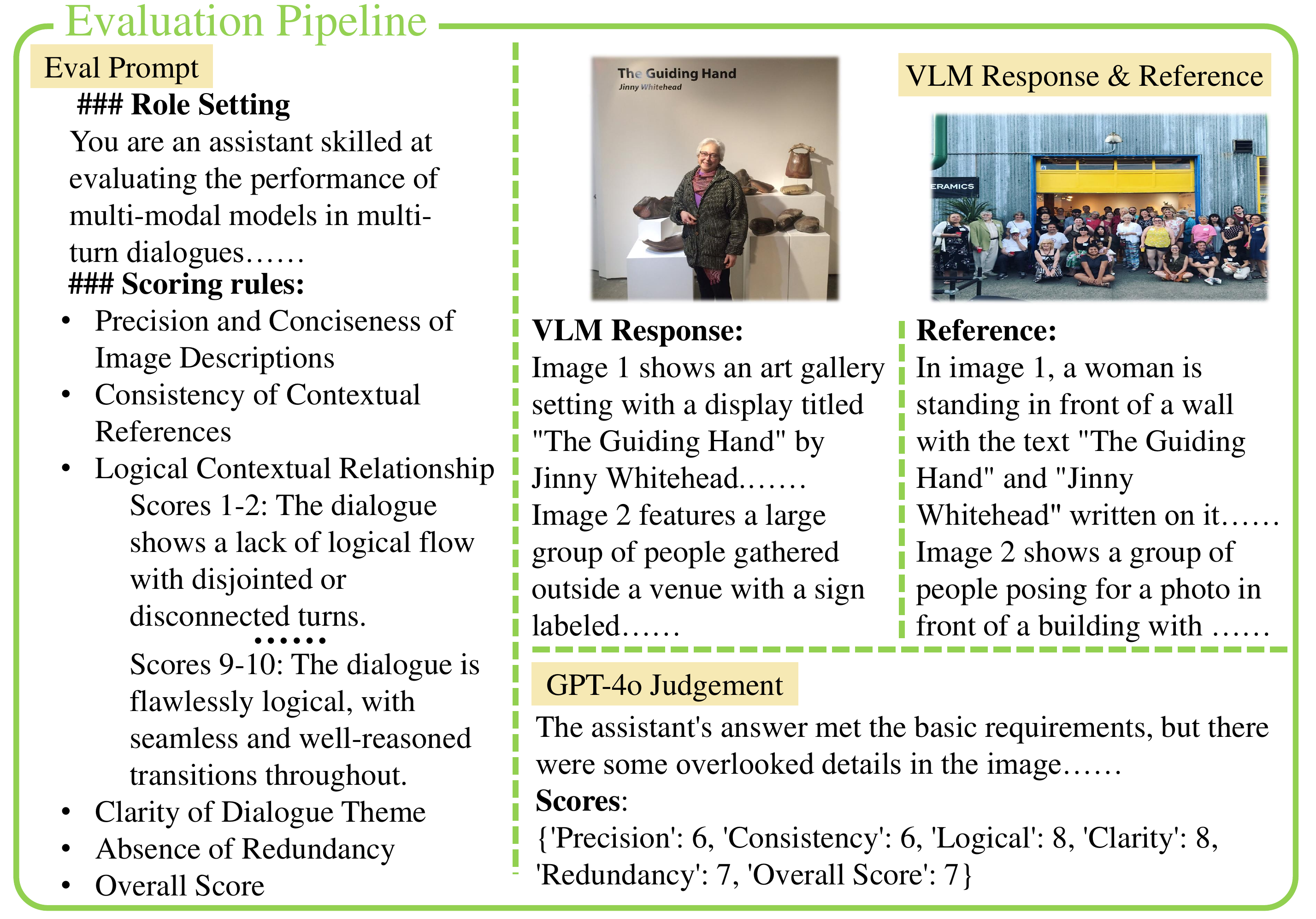}
    \caption{MMCR Evaluation Pipeline Diagram. We provide a prompt, the model's response, and the reference text, and then let GPT-4o serve as an impartial judge to score the outputs.}
    \label{fig:eval_prompt}
\end{figure}

\noindent \textbf{Evaluation.}
Using a powerful LLM as an impartial judge to evaluate model responses has always been a common paradigm in both the NLP and VLM fields. This approach not only avoids the subjective differences introduced by individual human judges but also provides relatively objective assessments for long texts\cite{zheng2023judgingllmasajudgemtbenchchatbot}. Inspired by MMDU, we have similarly developed an evaluation pipeline using GPT-4o to assess model performance. We combine the model's inference results on the proposed MMCR-Bench with specific prompts and the corresponding reference answers from the benchmark, then submit them to GPT-4o to score the model’s performance. Fig.~\ref{fig:eval_prompt} illustrates the prompt designed for contextual logic reasoning along with the corresponding question and response structure. Our evaluation focuses on the overall multi-turn dialogue. Unlike MMDU, which emphasizes creativity and richness, our goal is to enhance and assess the contextual reasoning abilities of VLMs in multi-turn dialogue scenarios. Therefore, we place greater emphasis on referential consistency, logical coherence, and dialogues that revolve around a clear topic. Accordingly, we have established five evaluation criteria: precision and conciseness of image descriptions, consistency of contextual references, logical contextual relationships, clarity of dialogue topics, degree of redundancy, and over all score. We guide GPT-4o to score across these dimensions using prompts that divide the range from 0 to 10 into two-point intervals corresponding to different scoring rules. We then aggregate and average the model's scores on the entire MMCR-Bench to obtain a final overall score.
Through this evaluation pipeline, we can obtain a comprehensive analysis of the model's contextual reasoning capabilities.

\section{Experiments}

In this section, we present the effectiveness of the proposed MMCR on both MMCR-Bench and public evaluation benchmarks. In Section~\ref{sec:mmcr_eval}, we report the evaluation results of SoTA VLMs for contextual reasoning using MMCR-Bench. In Section~\ref{sec:public_eval}, we provide comparative results from incorporating MMCR data during the instruction fine-tuning phase of VLM models.

\begin{table*}[]
\centering
  \resizebox{1\textwidth}{!}{%
\begin{tabular}{c|c|c|ccccccc}
\toprule
Model                     & LLM                                     & ViT                                       & MMCR-310k              & PC                          & CC                          & LC                          & CT                          & DR                          & OA                          \\ \hline
                          &                                         &                                           & \ding{56} & 57.9                        & 65.9                        & 58.4                        & 68.8                        & 61.8                        & 60.6                        \\
                          &                                         &                                           & \ding{52} & 61.3                        & 68.4                        & 62.0                        & 71.2                        & 69.3                        & 65.8                        \\ \cline{4-10} 
\multirow{-3}{*}{Ovis-1B} & \multirow{-3}{*}{Qwen2.5-0.5B-Instruct} & \multirow{-3}{*}{aimv2-large-patch14-448} & $\Delta$                     & {\color[HTML]{0ADF0B} +3.4} & {\color[HTML]{0ADF0B} +2.5} & {\color[HTML]{0ADF0B} +3.6} & {\color[HTML]{0ADF0B} +2.4} & {\color[HTML]{0ADF0B} +7.5} & {\color[HTML]{0ADF0B} +5.2} \\ \midrule
                          &                                         &                                           & \ding{56} & 62.6                        & 72.8                        & 67.3                        & 75.7                        & 68.9                        & 68.5                        \\
                          &                                         &                                           & \ding{52} & 65.8                        & 74.6                        & 69.9                        & 76.8                        & 75.0                        & 72.0                        \\ \cline{4-10} 
\multirow{-3}{*}{Ovis-4B} & \multirow{-3}{*}{Qwen2.5-3B-Instruct}   & \multirow{-3}{*}{aimv2-huge-patch14-448}  & $\Delta$                     & {\color[HTML]{0ADF0B} +3.2} & {\color[HTML]{0ADF0B} +1.8} & {\color[HTML]{0ADF0B} +2.6} & {\color[HTML]{0ADF0B} +1.1} & {\color[HTML]{0ADF0B} +6.1} & {\color[HTML]{0ADF0B} +3.5} \\ \midrule
                          &                                         &                                           & \ding{56} & 63.2                        & 75.1                        & 68.7                        & 77.1                        & 72.7                        & 70.4                        \\
                          &                                         &                                           & \ding{52} & 67.0                        & 76.5                        & 72.5                        & 79.1                        & 76.6                        & 74.6                        \\ \cline{4-10} 
\multirow{-3}{*}{Ovis-8B} & \multirow{-3}{*}{Qwen2.5-7B-Instruct}   & \multirow{-3}{*}{aimv2-huge-patch14-448}  & $\Delta$                     & {\color[HTML]{0ADF0B} +3.8} & {\color[HTML]{0ADF0B} +1.4} & {\color[HTML]{0ADF0B} +3.8} & {\color[HTML]{0ADF0B} +2.0} & {\color[HTML]{0ADF0B} +3.9} & {\color[HTML]{0ADF0B} +4.2} \\ \bottomrule
\end{tabular}
}
\caption{Evaluation results of Ovis on MMCR-Bench. We use Qwen2.5-Instruct~\cite{qwen2025qwen25technicalreport} and aimv2~\cite{fini2024multimodalautoregressivepretraininglarge} as the model's LLM and visual encoder, respectively. We report the metrics of precision and conciseness (PC) of image descriptions , consistency of contextual (CC) references, logical contextual (LC) relationships, clarity of dialogue topics (CT), degree of redundancy (DR), and over all score (OA).}
\label{tab:mmcr_eval}
\end{table*}

\begin{table*}[t]
\centering
  \resizebox{1\textwidth}{!}{%
\begin{tabular}{lcccccccccccc}
\toprule
\multicolumn{1}{l|}{Models}                                                      & \multicolumn{1}{c|}{Param (B)}                                              & \multicolumn{1}{c|}{LLM}                                                             & AI2D                        & HB                          & MV                          & MMB                         & MMMU                        & MMS                         & MMV                         & OCRB                        & SQA                         & MME                         \\ \hline
\multicolumn{13}{c}{\textit{Closed-source VLMs}}                                                                                                                   \\ \hline
\multicolumn{1}{l|}{{\color[HTML]{262626} GPT-4o (0513, detail-low)~\cite{openai2024gpt4o}}}            & \multicolumn{1}{c|}{{\color[HTML]{262626} -}}                                        & \multicolumn{1}{c|}{-}                                                             & {\color[HTML]{262626} 77.4} & {\color[HTML]{262626} 51.7} & {\color[HTML]{262626} 57.2} & {\color[HTML]{262626} 82.8} & {\color[HTML]{262626} 62.8} & {\color[HTML]{262626} 61.6} & {\color[HTML]{262626} 66.5} & {\color[HTML]{262626} 73.5} & {\color[HTML]{262626} 90.1} & {\color[HTML]{262626} 2329} \\
\multicolumn{1}{l|}{{\color[HTML]{262626} Gemini-1.5-Flash~\cite{geminiteam2024gemini15unlockingmultimodal}}}                     & \multicolumn{1}{c|}{{\color[HTML]{262626} -}}                                        & \multicolumn{1}{c|}{-}                                                             & {\color[HTML]{262626} 78.5} & {\color[HTML]{262626} 48.5} & {\color[HTML]{262626} 51.3} & {\color[HTML]{262626} 76.9} & {\color[HTML]{262626} 58.2} & {\color[HTML]{262626} 55.8} & {\color[HTML]{262626} 63.2} & {\color[HTML]{262626} 75.3} & {\color[HTML]{262626} 83.3} & {\color[HTML]{262626} 2078} \\
\multicolumn{1}{l|}{{\color[HTML]{262626} Claude3.5-Sonnet~\cite{claude2024}}}                     & \multicolumn{1}{c|}{{\color[HTML]{262626} -}}                                        & \multicolumn{1}{c|}{-}                                                             & {\color[HTML]{262626} 80.2} & {\color[HTML]{262626} 49.9} & {\color[HTML]{262626} 61.8} & {\color[HTML]{262626} 78.5} & {\color[HTML]{262626} 65.9} & {\color[HTML]{262626} 62.2} & {\color[HTML]{262626} 66.0} & {\color[HTML]{262626} 78.8} & {\color[HTML]{262626} 88.9} & {\color[HTML]{262626} 1920} \\ 
\hline
\multicolumn{13}{c}{\textit{Open-source VLMs}}                                                                                                                                                                                                                                                                                                                                                                                                                                                                                                                                                                                                     \\ \hline
\rowcolor[HTML]{EFEFEF} 
\multicolumn{1}{l|}{\cellcolor[HTML]{EFEFEF}{\color[HTML]{262626} SmolVLM-500M~\cite{marafioti2025smolvlm}}} & \multicolumn{1}{c|}{\cellcolor[HTML]{EFEFEF}{\color[HTML]{262626} 0.5}} & \multicolumn{1}{c|}{\cellcolor[HTML]{EFEFEF}{\color[HTML]{262626} SmolLM2-360M}}     & 59.2                        & 31.1                        & 39.8                        & 41.9                        & 33.6                        & 38.3                        & 25.7                        & 60.9                        & 80.0                        & 1395                        \\
\rowcolor[HTML]{EFEFEF} 
\multicolumn{1}{l|}{\cellcolor[HTML]{EFEFEF}{\color[HTML]{262626} H2OVL-800M~\cite{galib2024h2ovlmississippivisionlanguagemodels}}}   & \multicolumn{1}{c|}{\cellcolor[HTML]{EFEFEF}{\color[HTML]{262626} 0.8}} & \multicolumn{1}{c|}{\cellcolor[HTML]{EFEFEF}{\color[HTML]{262626} H2O-DANUBE3}}          & 53.5                        & 29.6                        & 39.8                        & 47.7                        & 32.1                        & 39.5                        & 30.2                        & 75.4                        & 69.8                        & 1469                        \\
\rowcolor[HTML]{EFEFEF} 
\multicolumn{1}{l|}{\cellcolor[HTML]{EFEFEF}LLaVA-OV-0.5B~\cite{li2024llavaonevisioneasyvisualtask}}                & \multicolumn{1}{c|}{\cellcolor[HTML]{EFEFEF}1}                          & \multicolumn{1}{c|}{\cellcolor[HTML]{EFEFEF}Qwen2-0.5B}                                          & 59.4                        & 27.9                        & 35.9                        & 56.8                        & 32.7                        & 37.7                        & 31.5                        & 58.3                        & 67.5                        & 1449                        \\
\rowcolor[HTML]{EFEFEF} 
\multicolumn{1}{l|}{\cellcolor[HTML]{EFEFEF}DeepSeek-VL-1.3B~\cite{lu2024deepseekvlrealworldvisionlanguageunderstanding}}                    & \multicolumn{1}{c|}{\cellcolor[HTML]{EFEFEF}2}                          & \multicolumn{1}{c|}{\cellcolor[HTML]{EFEFEF}DeekSeek-1B}                                 & 51.5                        & 27.6                        & 30.7                        & 63.8                        & 33.8                        & 39.9                        & 29.2                        & 41.3                        & 68.4                        & 1531                        \\
\rowcolor[HTML]{EFEFEF} 
\multicolumn{1}{l|}{\cellcolor[HTML]{EFEFEF}Janus-1.3B~\cite{wu2024janus}}                          & \multicolumn{1}{c|}{\cellcolor[HTML]{EFEFEF}2.1}                        & \multicolumn{1}{c|}{\cellcolor[HTML]{EFEFEF}DeepSeek-1B}                               & 52.8                        & 30.3                        & 33.7                        & 50.3                        & 31.2                        & 37.6                        & 37.5                        & 48.2                        & 75.1                        & 1579                        \\ \hline
\multicolumn{1}{l|}{Ovis-1B}                                                    & \multicolumn{1}{c|}{}                                                                & \multicolumn{1}{c|}{}                                                              & 74.9                        & 44.4                        & 56.7                        & 68.6                        & 35.4                        & \textbf{51.6}               & 47.6                        & 86.5                        & 81.1                        & 1664                        \\
\multicolumn{1}{l|}{Ovis-1B + MMCR}                                         & \multicolumn{1}{c|}{\multirow{-2}{*}{1}}                               & \multicolumn{1}{c|}{\multirow{-2}{*}{Qwen2.5-0.5B-Inst}}                                   & \textbf{76.0}               & \textbf{44.6}               & \textbf{56.9}               & \textbf{69.3}               & \textbf{36.6}               & 51.5                        & \textbf{48.8}               & \textbf{86.6}               & \textbf{81.6}               & \textbf{1689}               \\ \hline
\rowcolor[HTML]{EFEFEF} 
\multicolumn{1}{l|}{\cellcolor[HTML]{EFEFEF}Phi-3.5-Vision~\cite{abdin2024phi3technicalreporthighly}}                      & \multicolumn{1}{c|}{\cellcolor[HTML]{EFEFEF}4}                          & \multicolumn{1}{c|}{\cellcolor[HTML]{EFEFEF}Phi-3.5}                                 & 77.8                        & 40.5                        & 43.3                        & 67.4                        & 44.6                        & 47.5                        & 43.2                        & 59.9                        & 88.9                        & 1838                        \\
\rowcolor[HTML]{EFEFEF} 
\multicolumn{1}{l|}{\cellcolor[HTML]{EFEFEF}InternVL-Chat-4B-V1.5~\cite{chen2024internvl}}          & \multicolumn{1}{c|}{\cellcolor[HTML]{EFEFEF}4}                          & \multicolumn{1}{c|}{\cellcolor[HTML]{EFEFEF}Phi-3}                                        & 77.0                        & 43.0                        & 55.0                        & 69.7                        & 45.1                        & 53.1                        & 43.6                        & 63.9                        & 92.6                        & 2079                        \\
\rowcolor[HTML]{EFEFEF} 
\multicolumn{1}{l|}{\cellcolor[HTML]{EFEFEF}Vintern-3B-beta~\cite{doan2024vintern1befficientmultimodallarge}}                     & \multicolumn{1}{c|}{\cellcolor[HTML]{EFEFEF}3.7}                        & \multicolumn{1}{c|}{\cellcolor[HTML]{EFEFEF}Qwen-2.5-3B}                             & 69.1                        & 43.2                        & 43.6                        & 66.6                        & 46.7                        & 47.5                        & 37.8                        & 61.8                        & 75.0                        & 1783                        \\
\rowcolor[HTML]{EFEFEF} 
\multicolumn{1}{l|}{\cellcolor[HTML]{EFEFEF}XGen-MM-Inst-IL-v1.5~\cite{blip3-xgenmm}}    & \multicolumn{1}{c|}{\cellcolor[HTML]{EFEFEF}4.4}                        & \multicolumn{1}{c|}{\cellcolor[HTML]{EFEFEF}Phi-3}                                               & 74.2                        & 39.8                        & 40.6                        & 69.8                        & 40.9                        & 48.4                        & 40.2                        & 55.1                        & 88.3                        & 1809                        \\ \hline
\multicolumn{1}{l|}{Ovis-4B}                                                    & \multicolumn{1}{c|}{}                                                                & \multicolumn{1}{c|}{}                                                              & 84.4                        & 52.3                        & \textbf{66.6}               & 78.4                        & 49.6                        & 59.0                        & \textbf{62.8}               & 87.5                        & 92.9                        & 2113                        \\
\multicolumn{1}{l|}{Ovis-4B + MMCR}                                         & \multicolumn{1}{c|}{\multirow{-2}{*}{4}}                               & \multicolumn{1}{c|}{\multirow{-2}{*}{Qwen2.5-3B-Inst}}                                     & \textbf{84.6}               & \textbf{52.7}               & \textbf{66.6}               & \textbf{79.3}               & \textbf{50.4}               & \textbf{59.5}               & \textbf{62.8}               & \textbf{87.9}               & \textbf{93.0}               & \textbf{2132}               \\ \hline
\rowcolor[HTML]{EFEFEF} 
\multicolumn{1}{l|}{\cellcolor[HTML]{EFEFEF}Molmo-7B-D~\cite{deitke2024molmopixmoopenweights}}                          & \multicolumn{1}{c|}{\cellcolor[HTML]{EFEFEF}8}                          & \multicolumn{1}{c|}{\cellcolor[HTML]{EFEFEF}Qwen2-7B}                               & 79.6                        & 47.7                        & 48.7                        & 70.9                        & 48.7                        & 54.4                        & 53.3                        & 69.4                        & 92.2                        & 1784                        \\
\rowcolor[HTML]{EFEFEF} 
\multicolumn{1}{l|}{\cellcolor[HTML]{EFEFEF}Llama-3.2-11B-VI~\cite{grattafiori2024llama3herdmodels}}       & \multicolumn{1}{c|}{\cellcolor[HTML]{EFEFEF}11}                         & \multicolumn{1}{c|}{\cellcolor[HTML]{EFEFEF}Llama-3.1-8B}                               & 77.3                        & 40.3                        & 47.7                        & 65.8                        & 48.0                        & 49.8                        & 57.6                        & 75.3                        & 83.9                        & 1821                        \\
\rowcolor[HTML]{EFEFEF} 
\multicolumn{1}{l|}{\cellcolor[HTML]{EFEFEF}Pixtral-12B~\cite{agrawal2024pixtral12b}}                         & \multicolumn{1}{c|}{\cellcolor[HTML]{EFEFEF}13}                         & \multicolumn{1}{c|}{\cellcolor[HTML]{EFEFEF}Nemo-12B}                                & 79.0                        & 47.0                        & 56.3                        & 72.7                        & 51.1                        & 54.5                        & 58.5                        & 68.5                        & 87.2                        & 1922                        \\
\rowcolor[HTML]{EFEFEF} 
\multicolumn{1}{l|}{\cellcolor[HTML]{EFEFEF}WeMM}                                & \multicolumn{1}{c|}{\cellcolor[HTML]{EFEFEF}7}                          & \multicolumn{1}{c|}{\cellcolor[HTML]{EFEFEF}InternLM2-7B}                             & 77.9                        & 47.5                        & 54.9                        & 75.7                        & 45.3                        & 57.0                        & 45.0                        & 62.8                        & 83.3                        & 2150                        \\ \hline
\multicolumn{1}{l|}{Ovis-8B}                                                    & \multicolumn{1}{c|}{}                                                                & \multicolumn{1}{c|}{}                                                              & 86.4                        & 55.9                        & 67.6                        & \textbf{80.2}               & 56.2                        & 63.1                        & 67.2                        & 87.1                        & 94.2                        & 2164                        \\
\multicolumn{1}{l|}{Ovis-8B + MMCR}                                         & \multicolumn{1}{c|}{\multirow{-2}{*}{8}}                               & \multicolumn{1}{c|}{\multirow{-2}{*}{Qwen2.5-7B-Inst}}                            & \textbf{86.6}               & \textbf{56.3}               & \textbf{68.0}               & 79.8                        & \textbf{57.3}               & \textbf{63.2}               & \textbf{67.6}               & \textbf{87.3}               & \textbf{94.6}               & \textbf{2237}               \\ \toprule
\end{tabular}
}
\caption{Evaluation results for multiple models on various public benchmarks. We use the scores from AI2D~\cite{hiippala2021ai2d}, HallusionBench (HB)~\cite{guan2024hallusionbench}, MathVista (MV)~\cite{lu2023mathvista}, MMBench-TEST-EN (MMB)~\cite{liu2023mmbench}, MMMU-VAL (MMMU)~\cite{yue2024mmmu}, MMStar (MS)~\cite{chen2024we}, MMVet (MMV)~\cite{yu2023mm}, OCRBench (OCRB)~\cite{liu2023hidden}, ScienceQA (SQA)~\cite{saikh2022scienceqa}, and MME~\cite{fu2024mmecomprehensiveevaluationbenchmark} as evaluation metrics. The results of others are sourced from the official OpenCompass publicly available leaderboard~\cite{duan2024vlmevalkit}. The best results are \textbf{bold}.}
\label{tab:public_eval}
\end{table*}

\subsection{Evaluation of multi-turn dialogue contextual reasoning}
\label{sec:mmcr_eval}
Tab.~\ref{tab:mmcr_eval} presents the results on MMCR-Bench on targeted dimensions. We employ three versions (1B, 4B, and 8B) of Ovis~\cite{lu2024ovis}, a novel VLM architecture designed to align visual and textual embeddings via a visual embedding table, as VLMs for our experiments. Performance comparison with other open-source and closed-source models in Tab.~\ref{tab:public_eval} confirms its SOTA performance. According to the experimental results, we can draw several observations:
\begin{itemize}
    \item Challenging Evaluation Environment: Even the current leading SOTA models face significant challenges under the MMCR-Bench evaluation framework. For example, Ovis-4B, as the current top open-source model, achieves results predominantly around 70\%, indicating that there is still considerable room for improvement in the contextual reasoning abilities of current VLMs.
    \item Quality of the MMCR Fine-Tuning Dataset: The MMCR instruction fine-tuning dataset yields average improvements of 4.1\%, 3.1\%, and 3.2\% across six evaluation metrics for the three groups of VLMs, demonstrating its excellent quality. Notably, the Ovis-1B model shows an overall improvement of 5.2\%, which further validates the contribution and beneficial impact of the proposed dataset on enhancing the contextual reasoning capabilities of multimodal models.
    \item Mitigating Redundancy in Smaller Models: Interestingly, under the Absence of Redundancy evaluation metric, we observe that smaller models exhibit greater improvements after instruction fine-tuning with MMCR. This suggests that redundancy issues are more pronounced in smaller models. Moreover, significant score enhancements in the corresponding metric post fine-tuning prove that our MMCR instruction fine-tuning dataset is highly effective in mitigating model redundancy.
\end{itemize}

\subsection{Comparison and improvement on SOTA}
\label{sec:public_eval}
In Tab.~\ref{tab:public_eval}, we present the performance of a range of recently released open-source and closed-source models across multiple public evaluation benchmarks. We group the open-source models and the various-sized Ovis models by their parameter counts into three groups, and incorporate MMCR data into the instruction fine-tuning phase of the Ovis models. This setup not only demonstrates the powerful multimodal capabilities of our baseline Ovis model, but also shows that our proposed MMCR instruction fine-tuning data yields considerable performance improvements. The results indicate that MMCR data can bring noticeable enhancements on existing public benchmarks. For instance, with the Ovis-1B model, the introduction of MMCR leads to improvements of +1.1\% on AI2D, +1.2\% on MMMU and MMVet, and an overall boost of 0.6\% across a total of 10 datasets. These enhancements underscore the contribution of our MMCR dataset to the performance of VLMs on general multimodal tasks.

\begin{table}[t]
\centering
  \resizebox{0.5\textwidth}{!}{%
\begin{tabular}{c|c@{\,}c@{\,}c@{\,}c@{\,}c@{\,}c@{\,}c@{\,}c@{\,}c@{\,}c@{\,}}
\hline
Model Ovis-1B & AI2D          & HB            & MV            & MMB           & MMMU          & MMS           & MMV           & OCRB          & SQA           & MME           \\ \hline
base          & 74.9          & \underline{44.4}    & \underline{56.7}    & \underline{68.6}    & 35.4          & \textbf{51.6} & 47.6          & \underline{86.5}    & 81.1          & 1664          \\
MMDU          & \underline{75.1}    & 44.2          & 55.9          & 67.8          & \underline{35.9}    & 51.3          & \textbf{50.9} & \textbf{86.6} & \underline{81.5}    & \underline{1667}    \\
MMCR          & \textbf{76.0} & \textbf{44.6} & \textbf{56.9} & \textbf{69.3} & \textbf{36.6} & \underline{51.5}    & \underline{48.8}    & \textbf{86.6} & \textbf{81.6} & \textbf{1689} \\ \hline
\end{tabular}
}
\caption{Comparison with baseline and MMDU on Ovis-1B. The best results are \textbf{bold} and the second-best results are \underline{underlined}.}
\label{tab:apple_compare}
\end{table}



\subsection{Apples-to-Apples Comparison}
To further demonstrate the superior quality of our proposed MMCR dataset, we conducted a performance comparison on the same baseline with another same type dataset MMDU \cite{liu2025mmdu} under approximately equal data volumes (50k for MMCR, 45k for MMDU). As shown in Tab.~\ref{tab:apple_compare}, both MMDU and MMCR improve performance compared to the baseline; however, our proposed MMCR delivers even greater gains, further validating the quality of MMCR.


\begin{figure}[t]
    \centering
    \includegraphics[scale=0.48]{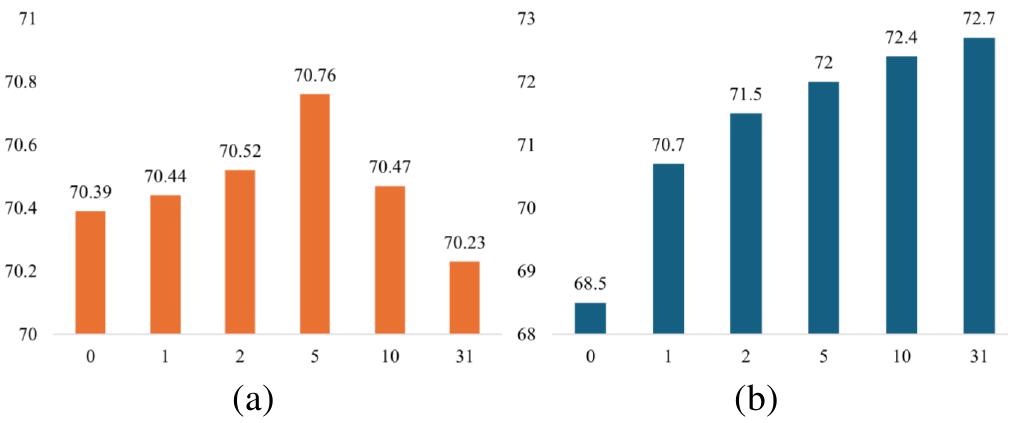}
    \caption{Histogram of Data Volume vs. Corresponding Metrics. (a) Average score across 9 public evaluation benchmarks for Ovis-4B at different data volumes. (b) OA score on MMCR-Bench for Ovis-4B at different data volumes. X-axis (unit: 10k).}
    \label{fig:less}
\end{figure}

\subsection{Biased Phenomenon: Less is More}
The fine-tuning phase of VLMs is critical for the model's final response performance. It is a widely accepted notion that more data leads to better performance, prompting researchers to continually incorporate larger volumes of data during the supervised fine-tuning stage. However, one aspect is often overlooked: when pursuing larger datasets, it is equally crucial to design a fine-tuning dataset with broad coverage and an even distribution.

As shown in Fig.~\ref{fig:less}, our experiments reveal that during the single instruction fine-tuning phase, continuously increasing the amount of MMCR data results in an upward trend in the average performance on both public evaluation benchmarks and MMCR-Bench. However, when the additional data reaches a certain proportion of the total, although MMCR-Bench still shows an upward trend, the performance on public evaluation benchmarks begins to decline. We believe this phenomenon is caused by the model developing a bias toward the corresponding data—improving performance in one direction while compromising overall performance.

Therefore, we conclude that ``Less is More":  using a smaller amount of data with a more rational distribution often achieves superior model performance. Moreover, maintaining a comprehensive and balanced data distribution during the fine-tuning phase is more important than merely expanding the dataset size.

\section{Conclusion}

In this paper, to enhance VLMs to meet the needs of daily human-AI conversations, we present MMCR, a multi-turn dialogue dataset involving multiple images. We thoroughly considered the characteristics of daily human conversations, such as multi-turn dialogues typically revolving around a few central topics, contextual referencing to maintain logical consistency. Through prompt engineering, we infused this information into GPT-4o to generate the data. Using the CLIP model, we filtered 1.6 million samples to obtain 310k high-quality multi-image, multi-turn dialogue samples. Experimental validation on the recently open-sourced Ovis show that fine-tuning with the proposed MMCR not only improves the metrics on MMCR-Bench but also further enhances the model's performance on public benchmarks. Finally, based on our experiments, we highlight a point often overlooked in training large models: more fine-tuning data is not always better, and maintaining a balanced proportion of data across different tasks is equally important.

{
    \small
    \bibliographystyle{ieeenat_fullname}
    \bibliography{main}
}

\end{document}